\documentclass[sn-mathphys]{sn-jnl}

\usepackage[export]{adjustbox}
\usepackage{booktabs} 
\usepackage{amsmath,amssymb}
\usepackage{graphicx}
\usepackage{mathtools}
\usepackage{color, soul}
\usepackage{bm}
\usepackage{siunitx}
\usepackage{multirow}
\usepackage{hyperref}
 \usepackage[normalem]{ulem}

\newcommand{\real}{\mathbb{R}}

\newcommand{\ie}{\emph{i.e.}}

\begin{document}

\title[Unsupervised OoD detection for safer retinal microsurgery]{Unsupervised out-of-distribution detection for safer robotically guided retinal microsurgery}


\author*[1]{\fnm{Alain} \sur{Jungo}}\email{alain.jungo@unibe.ch} 
\author[1]{\fnm{Lars} \sur{Doorenbos}}
\author[2]{\fnm{Tommaso} \sur{Da Col}}
\author[2]{\fnm{Maarten} \sur{Beelen}}
\author[3]{\fnm{Martin} \sur{Zinkernagel}}
\author[1]{\fnm{Pablo} \sur{Márquez-Neila}}
\author[1]{\fnm{Raphael} \sur{Sznitman}}


\affil[1]{\orgname{ARTORG Center, University of Bern}, \city{Bern}, \country{Switzerland}}
\affil[2]{\orgname{Preceyes B.V.}, \city{Eindhoven}, \country{The Netherlands}}
\affil[3]{\orgdiv{Department of Ophthalmology and Department of Clinical Research}, \orgname{Bern University Hospital}, \city{Bern}, \country{Switzerland}}

\abstract{
\textbf{Purpose:}
A fundamental problem in designing safe machine learning systems is identifying when samples presented to a deployed model differ from those observed at training time. Detecting so-called out-of-distribution (OoD) samples is crucial in safety-critical applications such as robotically guided retinal microsurgery, where distances between the instrument and the retina are derived from sequences of 1D images that are acquired by an instrument-integrated optical coherence tomography (iiOCT) probe.

\textbf{Methods:}
This work investigates the feasibility of using an OoD detector to identify when images from the iiOCT probe are inappropriate for subsequent machine learning-based distance estimation. We show how a simple OoD detector based on the Mahalanobis distance can successfully reject corrupted samples coming from real-world ex vivo porcine eyes.

\textbf{Results:}
Our results demonstrate that the proposed approach can successfully detect OoD samples and help maintain the performance of the downstream task within reasonable levels. MahaAD outperformed a supervised approach trained on the same kind of corruptions and achieved the best performance in detecting OoD cases from a collection of iiOCT samples with real-world corruptions.

\textbf{Conclusion:}
The results indicate that detecting corrupted iiOCT data through OoD detection is feasible and does not need prior knowledge of possible corruptions. Consequently, MahaAD could aid in ensuring patient safety during robotically guided microsurgery by preventing deployed prediction models from estimating distances that put the patient at risk.
}


\maketitle


\section{Introduction}
Ensuring safe machine learning models is one of the key challenges for real-world medical systems. While the need for reliable models is highly important for image-based diagnostics with human-in-the-loop users, it is mission-critical when combined with medical robotic systems that tightly couple image-based sensing for augmented visualizations or automation.

\begin{figure}[b]
\centering
\includegraphics[width=\textwidth]{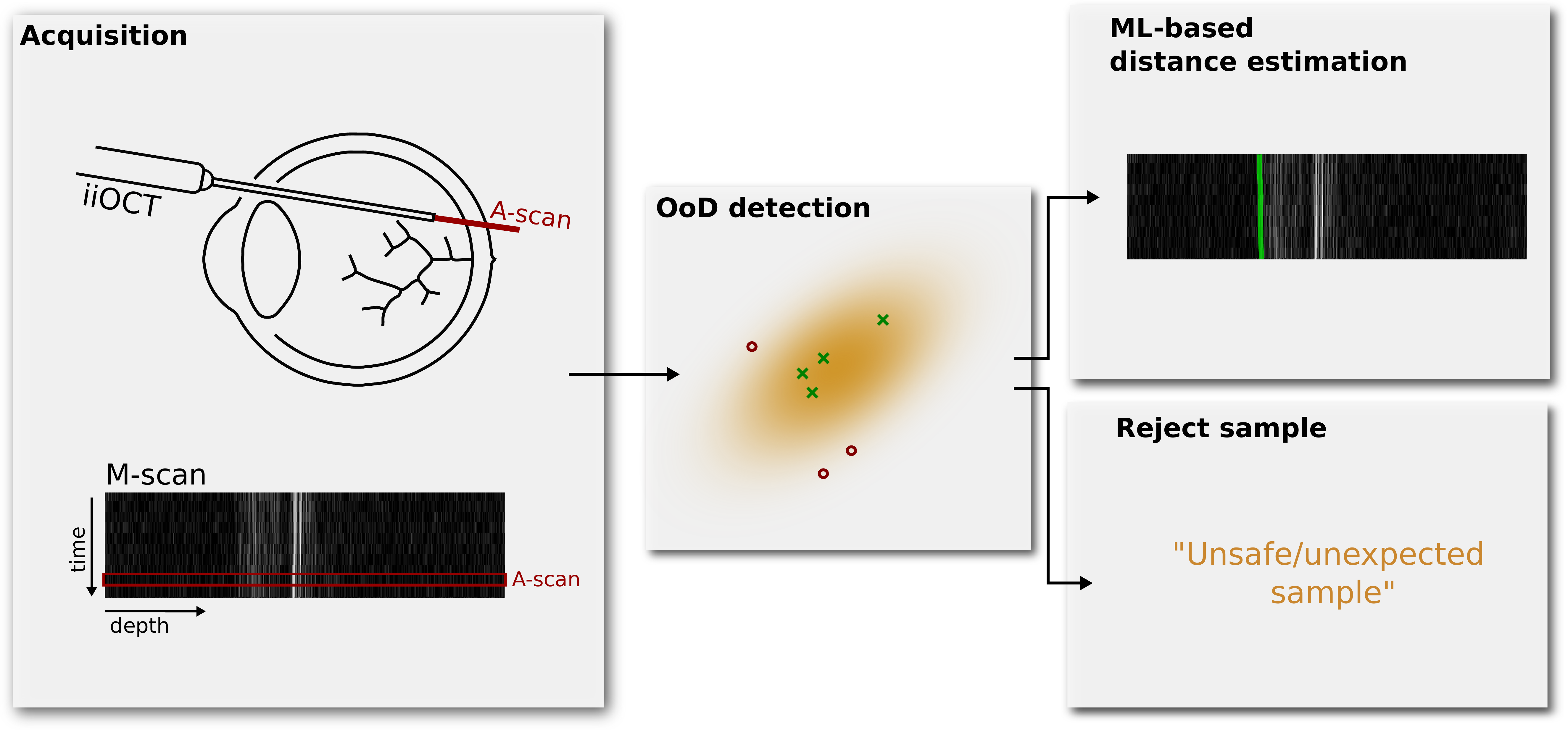}
\caption{Out-of-distribution detection of an inappropriate sequence of 1D~images, or {\it M-scan}, acquired by an iiOCT probe. These should be rejected rather than processed by a subsequent machine learning-based distance estimation method.}
\label{fig:concept}
\end{figure}

In this context, one of the fundamental problems in designing safe machine learning is identifying when samples presented to a deployed model differ from those observed at training time. This problem, commonly known as {\it out-of-distribution} (OoD) detection~\cite{yang2021generalized}, aims to alleviate the risks of evaluating OoD samples, as performances on these are known to be erratic and typically produce wrong answers with high confidences, whereby making them potentially dangerous. As machine learning has become increasingly prevalent in mission-critical systems, the problem of OoD detection has gathered significant attention both in general computer vision research~\cite{yang2021generalized}, and in applied medical imaging systems~\cite{marquez2019image,MaierHein2019,jungo2018uncertainty,zimmerer2022mood,schlegl2017unsupervised,berger2021confidence,gonzalez2022distance}.

OoD detection for robotically assisted surgery is particularly relevant as erratic machine learning predictions can have extremely serious consequences for the patient. For example, a misprediction in the distance estimation between an instrument and its targeted tissue could lead to important inadvertent trauma. Surprisingly, the topic of OoD detection for robotically assisted surgery has received little attention to date, despite its necessity and advantages. More broadly, the potential benefits of OoD detection in this context remain largely unexplored. This work aims to close this gap by analyzing the implications of integrating an OoD detector in a relevant robot-assisted surgery use case.

Specifically, we consider the setting of retinal microsurgery, where a machine learning model is needed to infer the distance between a robotically manipulated instrument and the retina of the eye (see Fig.~\ref{fig:concept}). As with most of the recently proposed robotic systems for retinal microsurgery~\cite{Balicki09,Uneri10,Poorten20,Cereda21,Weiss18}, the goal is to assist an operating surgeon when manipulating micron-sized retinal structures using an \emph{optical coherence tomography} (OCT) imaging probe which yields 1D OCT measures over time, also known as \emph{M-scans}. When using such a probe to help guide the robot to an intra-retinal injection site, automatic estimation between the instrument and the retinal surface is key. Yet, for any robot-tissue interacting system, a critical necessity is to ensure that the inferred distances derived from the imaging probe are safe for the robotic system to use.

\begin{figure}[t]
\centering
\includegraphics[width=1.0\textwidth]{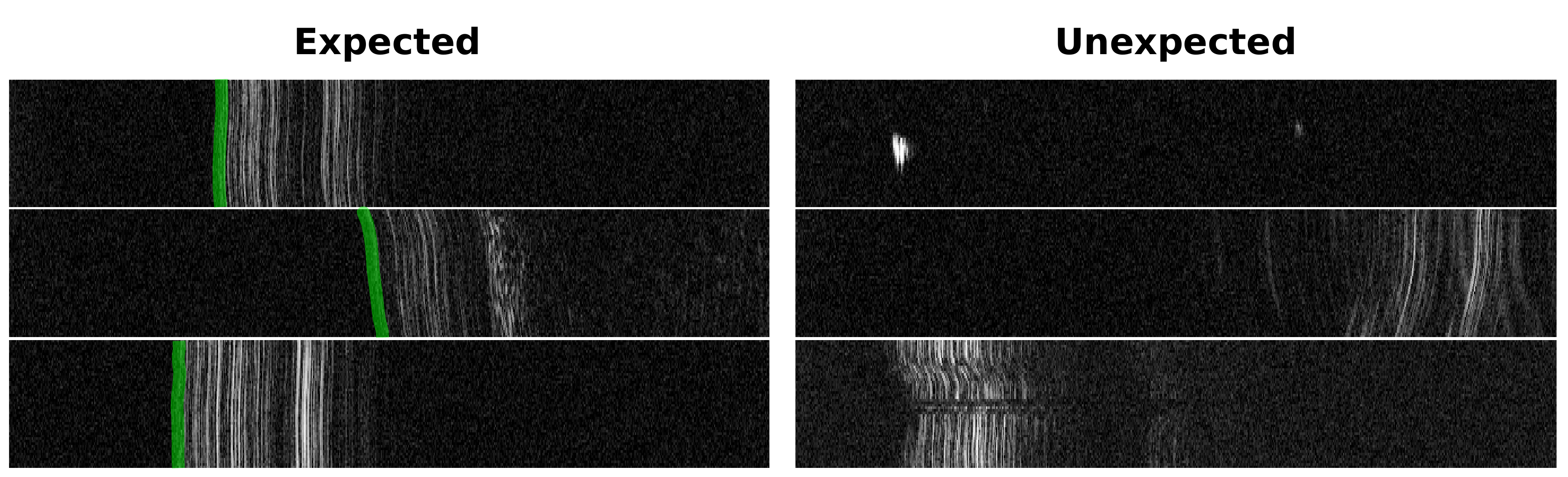}
\caption{Six M-scans acquired from a 1D OCT image probe from which distance estimates to the ILM of the retina (shown in green) need to be computed. Evaluating unexpected images (right column) can lead to incorrect estimates and endanger the intervention. Images were resized for improved visualization.}
\label{fig:motivation}
\end{figure}

To this end, this work investigates the feasibility of using OoD detection to identify when images from an \emph{intraoperative instrument-integrated OCT} (iiOCT) probe are inappropriate for subsequent machine learning-based distance estimation (see Fig.~\ref{fig:motivation}). We show how data from this probe, in combination with the simple MahaAD OoD~\cite{rippel2021modeling} detector, can be rejected from further evaluation when the data is corrupted. We demonstrate the implications of our approach on the downstream task of distance estimation using simulated corruptions and report OoD detection performance on ex vivo porcine eyes with real-world corruptions.

\section{Methods}

\subsection{Problem setting} 
Our retinal microsurgical setup is equipped with a robot that manipulates an injection needle with an iiOCT sensor. The sensor captures the retinal tissue in front of the instrument in a M-scan, which is a sequence of one-dimensional depth signals, denoted \emph{A-scans}. Specifically, {M}-scans contain useful information about the layers of the retina and the sensor's distance to the different layers~(see Fig.~\ref{fig:motivation}). However, extracting distance information from {M}-scans is challenging due to the large appearance variability and noise observed in these signals.

To this end, machine learning and deep learning models are a natural approach to do so consistently and reliably. We thus train a deep learning model~$r:\real^P\to[0,1]^P$ to estimate the location of the internal limiting membrane (ILM) of the retina. Given an M-scan $\mathbf{x}$, the retinal detection model~$r$ receives individual A-scans as one-dimensional vectors~$\mathbf{x}_j$ and produces one-dimensional heatmaps~$\hat{\mathbf{y}}_j=r(\mathbf{x}_{j})$ indicating the probability that the ILM is located at each location of the input A-scan. The location of maximum probability determines the ML-based distance as shown in Fig.~\ref{fig:concept}. Similar to~\cite{Lee2021}, the model~$r$ is trained by minimizing the standard L2 loss over a training dataset~$\mathcal{T}=\{(\mathbf{x}^{(i)}, \mathbf{y}^{(i)})\}_{i=1}^N$ of {M}-scans and their corresponding ground-truth retinal maps.

At inference time, the retinal detection model~$r$ is robust to the types of {A}-scan variability learned from the training set~$\mathcal{T}$, but not to others never seen in this dataset. This poses a risk to the safety of the surgical system in practice, as we cannot ensure that the range of potential perturbations that can occur during surgery are present in the training dataset. The range is simply too large to build a representative dataset that covers all cases.

\subsection{Unsupervised OoD detection}
\label{sec:method}
We augment our system with an unsupervised out-of-distribution detection method to tackle the abovementioned limitation. Our approach is unsupervised in the sense that we do not have examples of OoD cases from which we can train a supervised model to perform OoD. Instead, we have only the dataset from which the distance estimation model, $r$, is trained. In this context, we leverage the MahaAD method proposed by Rippel~\emph{et al.}~\cite{rippel2021modeling} to learn the appearance of {M}-scans in the training dataset and detect when novel {M}-scans are too far from the training distribution to be safely processed by~$r$. We select this model as it has been shown to be highly effective in a large number of cases while being interpretable and computationally lean~\cite{Doorenbos2022}. 

At training time, MahaAD learns the training distribution by fitting multiple multivariate Gaussians to latent representations of the training data {at different scales}. More specifically, we build a training dataset {$\mathcal{T}'=\{\mathbf{x}^{(i)}\}_{i=1}^{M}$, where each sample~$\mathbf{x}^{(i)}\in\real^{10\times{}P}$ is a M-scan of 10~consecutive A-scans. M-scans} in~$\mathcal{T}'$ may come from the training data~$\mathcal{T}$ used to train~$r$ or, given the unsupervised nature of MahaAD, from any other dataset of {M}-scans without annotations. Given a pre-trained network~$f$ with $K$~layers for feature extraction, MahaAD first describes each training sample~{$\mathbf{x}^{(i)}$} as a collection of $K$~feature vectors {$\{\mathbf{f}_{i,k}=f_k(\mathbf{x}^{(i)})\}_{k=1}^K$}, where each vector~{$f_k(\mathbf{x}^{(i)})$} is the spatial average of the $k$-th feature map for the input~{$\mathbf{x}^{(i)}$}. The collection of the feature vectors for all training samples is then used to fit $K$~multivariate Gaussians, one per layer~$k$, with parameters,
\[
\bm{\mu}_k=\frac{1}{N}\sum_{i=1}^{N}{\mathbf{f}_{i,k}} \quad\text{and}\quad \bm{\Sigma}_k=\frac{1}{N}\sum_{i=1}^{N}{(\mathbf{f}_{i,k} - \bm{\mu}_k)(\mathbf{f}_{i,k} - \bm{\mu}_k)^T}\quad\forall k\in\{1,\ldots,K\}.
\]

\begin{figure}[b]
\centering
\includegraphics[width=0.75\textwidth]{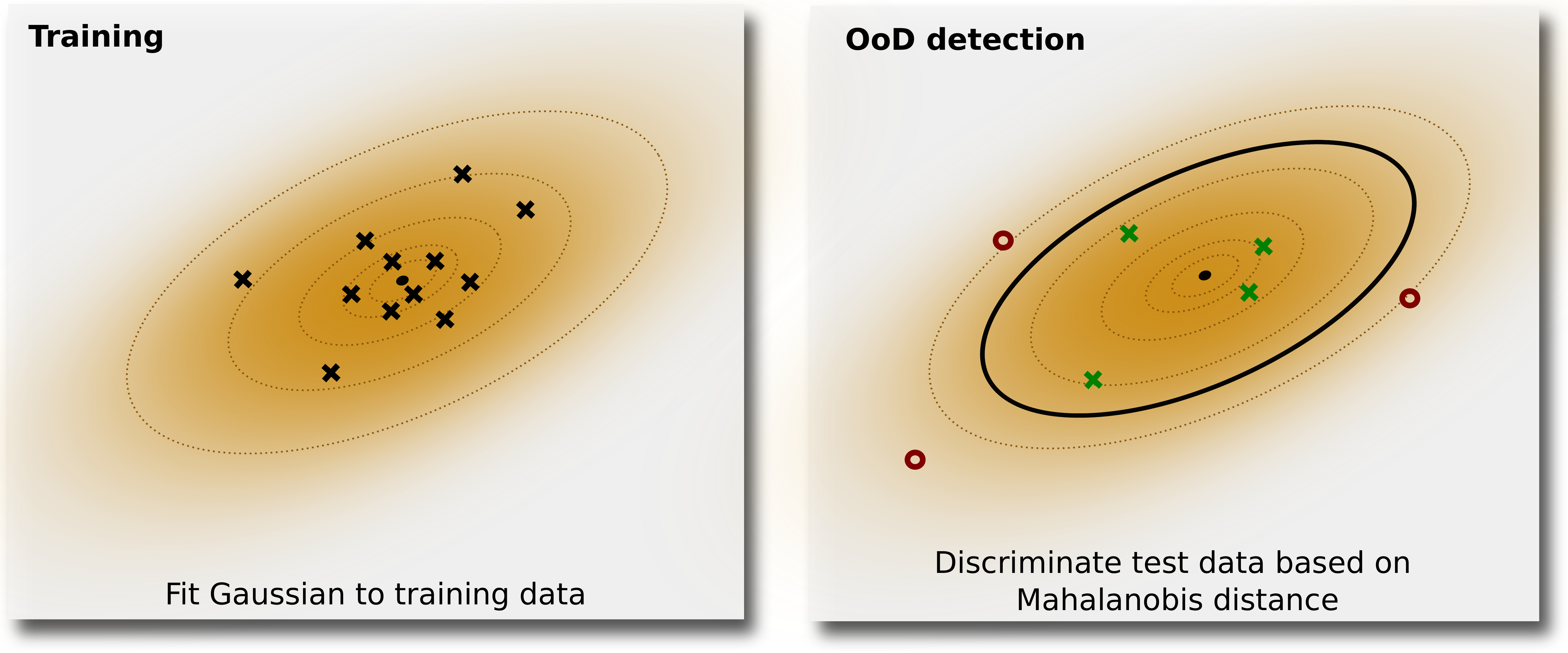}
\caption{Example of MahaAD with a single bivariate Gaussian (\ie,~a single 2D latent representation). A multivariate Gaussian is fit to the latent representations of the training samples and is used to determine the Mahalanobis distance of the test samples' latent representations. Based on the distance, samples are considered in- or out-of-distribution.}
\label{fig:maha_principle}
\end{figure}

At test time, MahaAD computes $K$~Mahalanobis distances between {an M-scan~$\mathbf{x}$} and the means~$\bm{\mu}_k$ of the learned Gaussians as shown in Figure~\ref{fig:maha_principle},
\[
d_k(\mathbf{x})=d(\x,\bm{\mu}_k)=\sqrt{(\mathbf{f}_k - \bm{\mu}_k)^T\bm{\Sigma_k}^{-1}(\mathbf{f}_k - \bm{\mu}_k)},\quad\forall k\in\{1,\ldots,K\}.
\]
The final OoD score for a test-time sample~{$\mathbf{x}$} is then the sum over all distances,
\[
s(\mathbf{x})=\sum_{k=1}^{K}{d_k}(\mathbf{x})\,.
\]
The {M-scan} is then considered OoD if its score {$s(\mathbf{x})$}~is larger than a threshold~$\tau$, which is the only hyperparameter of the method. When an M-scan is considered OoD, we treat all of its individual A-scan components {$\mathbf{x}_j$} as OoD and assume they are not suitable for safe estimation with the subsequent retina detection model~$r$. We experimentally found that applying MahaAD on {M-scans} produced more reliable results than on individual A-scans.

\section{Experimental setting}

\subsection{Data}\label{sec:data}
Our data consists of four recordings from ex vivo trials on four different pig eyes, with each recording containing approximately 900'000~A-scans. The iiOCT device produced temporal A-scans at a frequency of approximately 700Hz with a resolution of \SI{3.7}{\micro\meter}/pixel and a scan depth of $P=674$ pixels~(\SI{2.49}{\milli\meter}). Of the four pig recordings, three were used for training (and validation), and the fourth recording was held out for evaluation.  From the training recordings, a collection of 334 in-distribution M-scans consisting of 10 A-scans was used to train the OoD detector. We manually selected samples with identifiable retina to ensure that they are in-distribution samples.

\subsection{Implementation details}
To measure the impact of OoD samples on the retinal model~$r$, we compared six OoD detection strategies and one reference baseline.

\begin{figure}[t]
    \centering
    \includegraphics[width=0.8\textwidth]{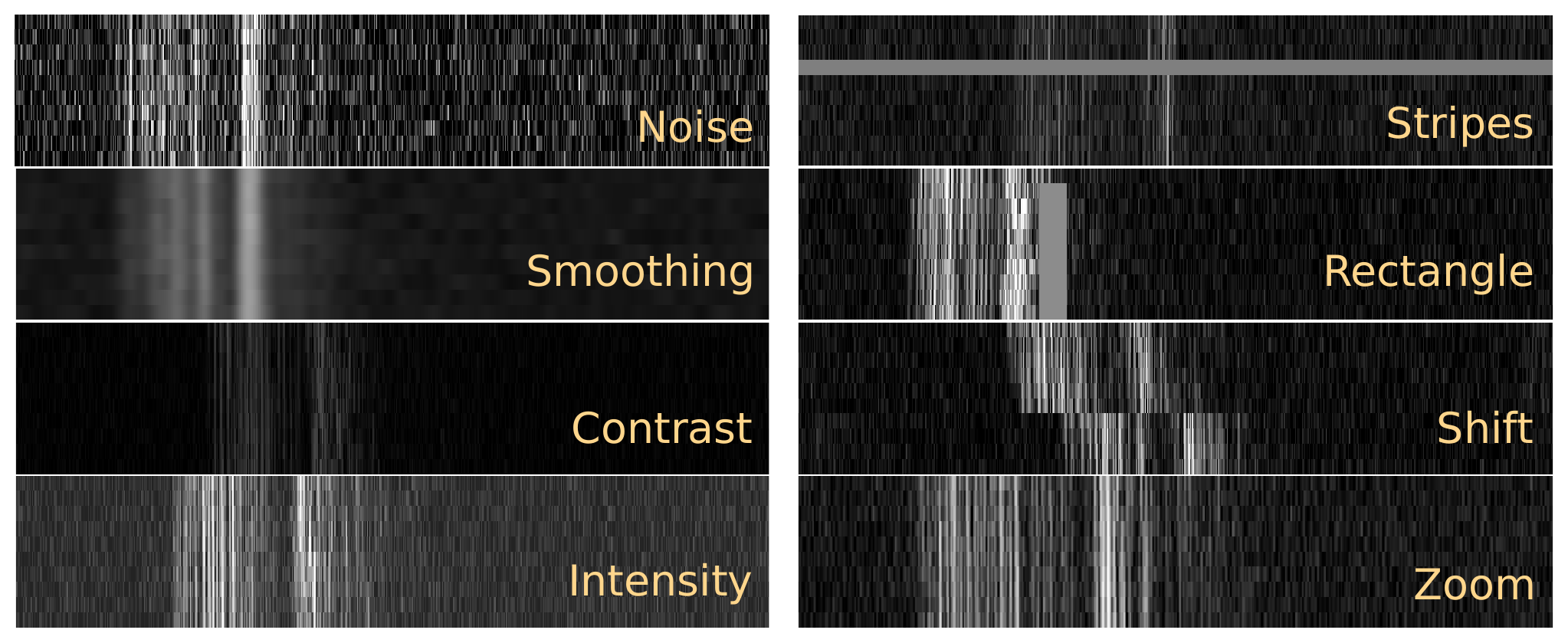}
    \caption{Examples of the eight types of perturbations applied to simulate OoD samples. Each sample {is an M-scan with a depth of 674 pixels and 10 consecutive A-scans}. Images were resized for improved visualization.}
    \label{fig:data}
\end{figure}

\renewcommand{\descriptionlabel}[1]{\hspace{\labelsep}\textbf{#1}}
\begin{description}
    \item[MahaAD:]
    The method proposed in Sect.~\ref{sec:method}. As proposed in \cite{rippel2021modeling}, our feature extractor~$f$ is an EfficientNet-B0~\cite{pmlr-v97-tan19a} with $K=9$~blocks pre-trained on ImageNet. The input to~$f$ are M-scans~$\mathbf{x}$ resized from $10\times{}674$ to $64\times{}224$ with bicubic interpolation to increase the importance of the temporal information and to make the input size closer to the training sizes of the EfficientNet-B0. We applied z-score normalization with the mean and standard deviation from ImageNet to all the input sequences.
    \item[Supervised:] An OoD detector implemented as a binary classifier and trained in a supervised fashion with both in-distribution and OoD samples. Given that OoD samples are not available in large amounts, we synthetically generated them by perturbing 50\% of the training data using four types of perturbations: noise, smoothing, shifts and intensity (see Fig.~\ref{fig:data}). The OoD detector uses an ImageNet-pre-trained EfficientNet-B0 as backbone with a classification head adapted to the binary OoD detection task. We fine-tuned all layers with Adam optimizer and learning rate $10^{-5}$.
    \item[Glow:] A generative flow-based model~\cite{kingma2018} used as OoD detector. We use the model's negative likelihood output as the OoD score (\ie, the lower the likelihood, the less probable a sample is in-distribution). The employed architecture has three blocks of 32 layers and was trained {with the public implementation of~\cite{amersfoort_glow_2022}}.
    \item[Uncertainty:] OoD~samples tend to produce estimations with lower maximum softmax probabilities, (\ie, higher uncertainty~\cite{hendrycks2016baseline}). We take the maximum probability of the estimated heatmap~$\hat{\mathbf{y}}_j=r(\mathbf{x}_j)$ and use its entropy as the OoD score.
    \item[Raw-MahaAD:] Similar to \textbf{MahaAD} but, instead of the feature vectors~$\mathbf{f}_{i,k}$, we use the raw signal to fit a single ($K=1$) multivariate Gaussian. This can be seen as an ablation of the deep features.
    \item[SNR:] A simple measure of scan quality directly used as OoD score. We measure the signal-to-noise ratio (SNR) as $\mu_{\mathbf{x}}/\sigma_{\mathbf{x}}$.
    \item[No-rejection]: Reference baseline that considers all samples as inliers (\ie, no OoD detection is applied).
\end{description}

In all cases, we used a retinal model~$r$ that was implemented as a one-dimensional U-Net-like architecture~\cite{ronneberger2015u} with four down-pooling/upsampling steps and one convolutional layer per step. We used Adam with a learning rate of $10^{-4}$ for optimization and performed early stopping according to the performance on the validation split. To train and validate~$r$, the location of the ILM was manually annotated for a random collection of 14'700 M-scans from the original pig recordings.

\subsection{Experiments}
\subsubsection{OoD detection for distance estimation} 
The first experiment measures the impact of our approach in a simulated scenario of retinal surgery where the retinal model~$r$ {only receives the samples considered safe for estimation by the OoD detector}. For this purpose, we employed a test set of 2'000 M-scans with annotated ILM locations. To account for the lack of real OoD~samples, OoD samples were synthetically generated by perturbing a fraction~$p$ of elements from the test data with eight types of corruptions: 

\renewcommand{\descriptionlabel}[1]{\hspace{\labelsep}\textit{#1}}
\begin{description}
    \item[Noise:] Additive Gaussian noise with $\mu$=0 and $\sigma$=50.
    \item[Smoothing:] Gaussian filtering with $\sigma$=5.
    \item[Contrast:] Contrast increase/decrease by a factor uniformly sampled from~$\{0.1, 0.2, 0.3, 2, 3, 4\}$.
    \item[Intensity:] Equally probable positive/negative shift of the intensity uniformly sampled from the set~$\mathcal{U}([-50,-25] \cup [25, 50])$.
    \item[Stripes:] Randomly replacing one or two A-scans in a sequence with a constant intensity sampled from~$\mathcal{U}(100,200)$.
    \item[Rectangle:] Randomly placing a rectangle with size {(\ie, M-scan stretch$\times{}$depth)} sampled from $\mathcal{U}([6,10]\times{}[15,30])$ pixels and a constant intensity sampled from~$\mathcal{U}(100,200)$.
    \item[Shift:] Roll in depth for a random split of A-scans in the sequence with positive/negative shift sampled from $\mathcal{U}(25,100)$~pixels.
    \item[Zoom:] Zoom each A-scan in a sequence by a factor sampled from~$\mathcal{U}(1.5,1.75)$.
\end{description}

All perturbations were applied with equal probability to samples with intensities rescaled to the range~$[0, 255]$. Figure~\ref{fig:data} shows examples of produced synthetic corruptions.

{\subsubsection{Real OoD samples in ex vivo data} 
In a second experiment, we explore the behavior of the methods when presented with real OoD {M-scans} that were manually identified in our data. For this purpose, we built a test dataset with 258~real OoD {M-scans} and 258~in-distribution {M-scans}, where each {M-scan} consists of 10~A-scans. Figure~\ref{fig:real_ex} includes a few examples. As these samples are real OoD cases, it is impossible to label the location of the ILM and thus prevents us from using the above experimental protocol. Instead, we compared the performance of the baselines in the task of detecting OoD samples in this small dataset, omitting the retinal network~$r$. }

\section{Results}

\subsection{OoD detection for distance estimation}

\begin{figure}[b]
\centering
\includegraphics[width=0.95\textwidth]{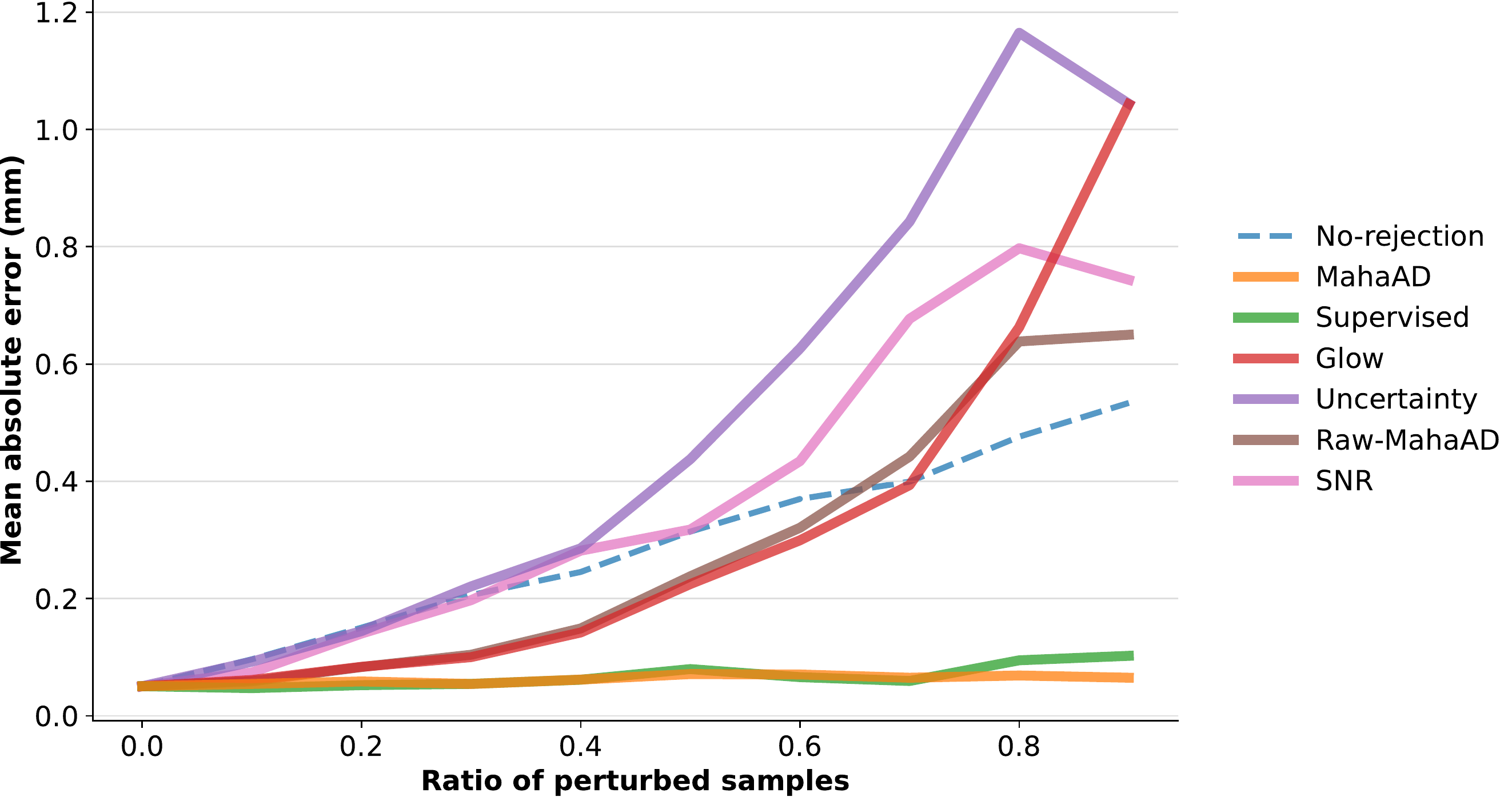}
\caption{Effect of different OoD methods on the retinal surgery pipeline. Mean absolute distance error (MAE) is shown for different perturbation ratios~$p$. For each baseline, a proportion of~$p$ {M-scans} were considered OoD and rejected from MAE computation.
}
\label{fig:reject}
\end{figure}

We measured the performance of~$r$ in terms of the mean absolute error~(MAE) between the estimated and the real distances for a progressively increasing ratio of corruptions~$p$, which ranged from~$0$~(\ie,~no corruptions) to~$0.9$. To quantify the impact of each OoD detection approach, M-scans detected as OoD were discarded from MAE computation. For proper comparison, an M-scan was considered OoD if it was among the top-$p$ highest OoD-scoring M-scans. Hence, a perfect OoD detector will discard all the corrupted M-scans, keeping the MAE low.

\textbf{MahaAD} outperformed all baselines, with its MAE staying almost constant for all the perturbation ratios~(Fig.~\ref{fig:reject}). \textbf{Raw-MahaAD}, \textbf{Glow}, \textbf{Uncertainty}, and \textbf{SNR} underperformed compared to \textbf{No-rejection}, suggesting that they flag a large proportion of correct samples as OoD while allowing corrupted A-scans to be processed by the subsequent retinal network. The poor behavior of \textbf{Uncertainty} and \textbf{SNR} is noticeable for perturbation ratios as low as~$0.2$, which makes them unsuitable for OoD detection in the present setting. Finally, \textbf{Supervised} matched the performance of \textbf{MahaAD}, but given that it was trained with the same kind of synthetic perturbations used in the evaluation, this is most likely an overoptimistic result.

\begin{figure}
\centering
\includegraphics[width=0.8\textwidth]{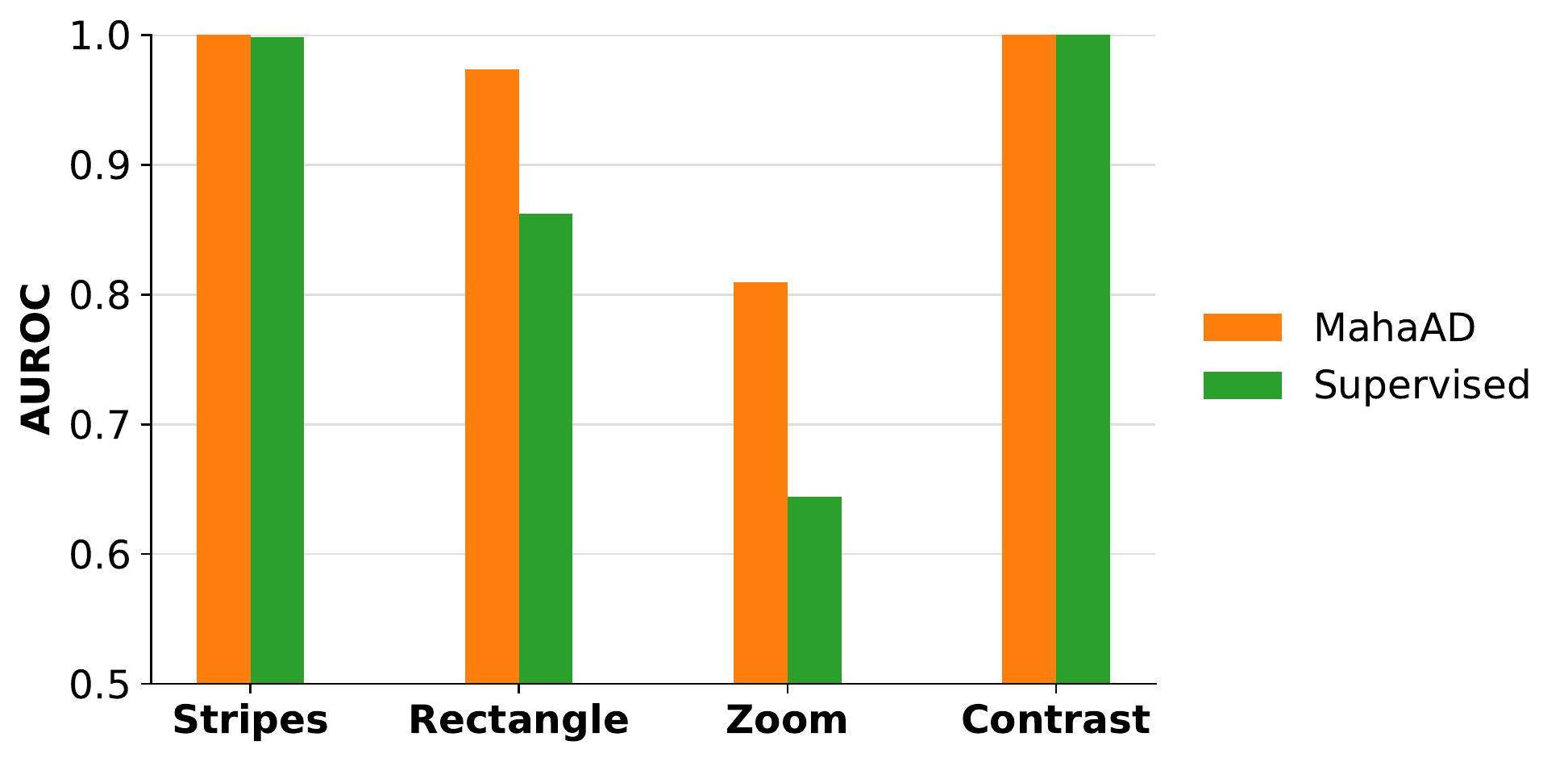}
\caption{{Comparison between \textbf{MahaAD} and \textbf{Supervised} on the OoD detection performance in terms of area under the receiver operating characteristic curve (AUROC) for corruptions not used for training \textbf{Supervised}.}}
\label{fig:pert_details}
\end{figure}

Additionally, we compared \textbf{MahaAD} and \textbf{Supervised} based on their isolated OoD detection performance for individual corruption types at a proportion~$p$ of~$0.5$. To investigate our presumption of \textbf{Supervised}'s overoptimistic performance due to known perturbations, we analyzed the corruptions that \textbf{Supervised} has not seen during training (\ie, \textit{stripes}, \textit{rectangle}, \textit{zoom}, \textit{contrast}). Figure~\ref{fig:pert_details} shows that \textbf{MahaAD} is outperforming \textbf{Supervised} in terms of OoD detection on the unseen corruptions. Specifically, the difference is notable for \textit{zoom} and \textit{rectangle}, which seem to be the most difficult perturbations to detect. This result indicates that \textbf{MahaAD} is a better OoD detector when the type of expected perturbations is unknown and for which we cannot train.

\subsection{Real OoD samples in ex vivo data} 
 
\begin{figure}[t]
\centering
\includegraphics[width=0.8\textwidth]{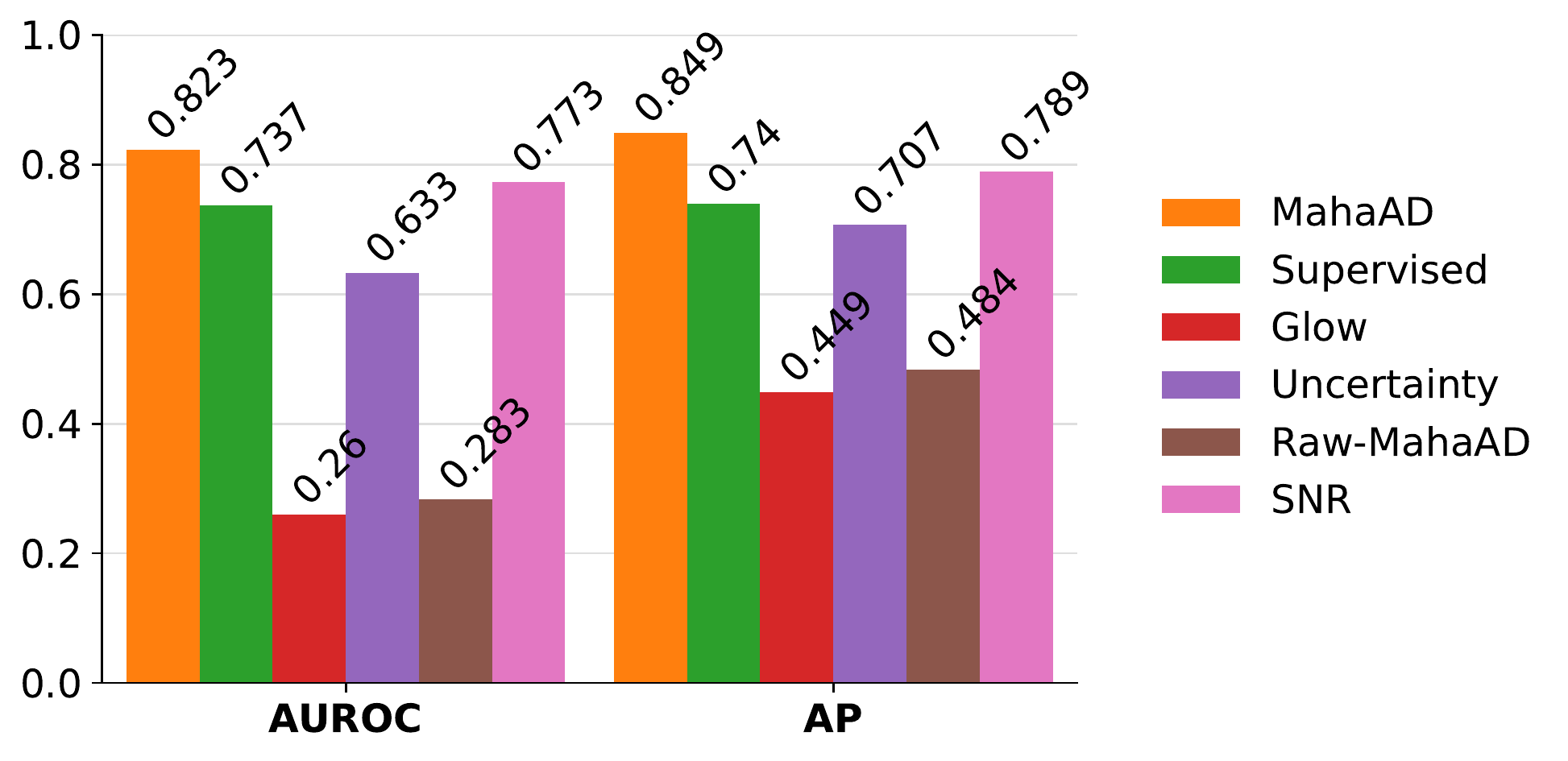}
\caption{{Area under the receiver operating characteristic curve (AUROC) and average precision (AP) performance for the detection task on real OoD samples.}}
\label{fig:real_ood}
\end{figure}

Figure~\ref{fig:real_ood} reports the results for the second experiment on the selection of real OoD samples. As previously found, \textbf{MahaAD} outperformed the other baselines, demonstrating its ability to generalize beyond simulated perturbations in a more realistic scenario. Furthermore, \textbf{Supervised} performs significantly worse than \textbf{MahaAD} on real data, suggesting that the results of Fig.~\ref{fig:reject} were indeed overoptimistic and that \textbf{Supervised} is not suitable as an OoD detector in a realistic scenario. In contrast, \textbf{SNR}'s performance improved on real data, likely due to a selection bias facilitating discrimination of OoD samples through low-order statistics. Surprisingly, \textbf{Glow} and \textbf{Raw-MahaAD} seem to produce OoD scores that better describe in-distribution samples than OoD samples.

\begin{figure}[b]
\centering
\includegraphics[width=1.0\textwidth]{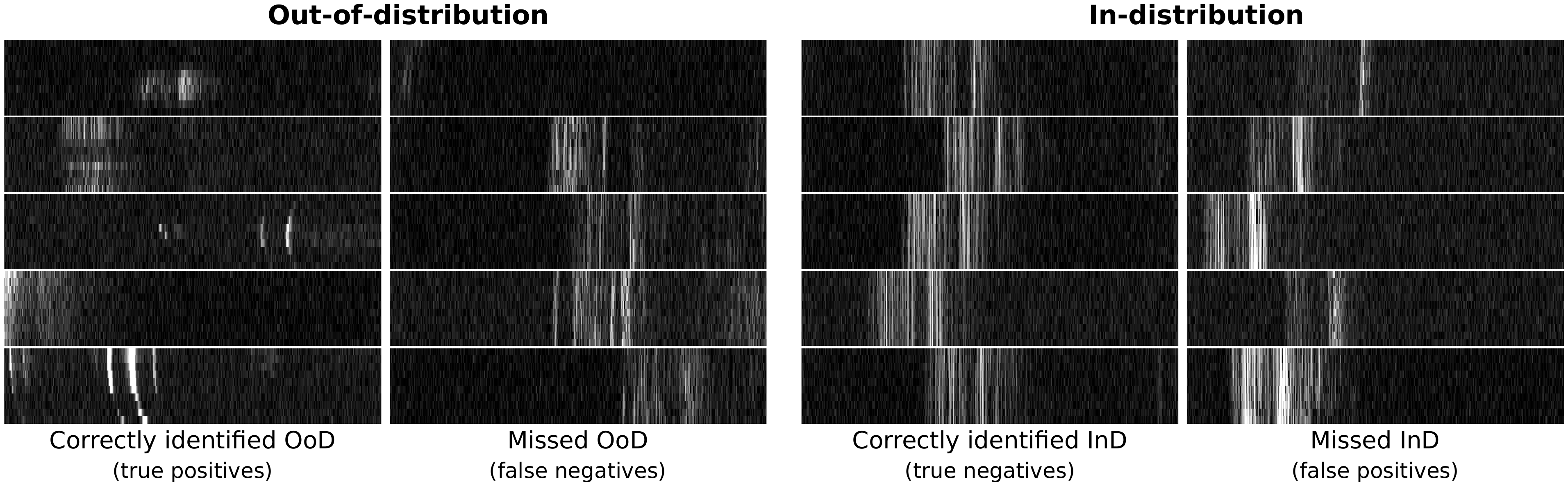}
\caption{Examples of correctly detected and missed OoD and in-distribution samples with \textbf{MahaAD}. Images have been resized for improved visualization.}
\label{fig:real_ex}
\end{figure}

Figure~\ref{fig:real_ex} shows visual examples of correctly and incorrectly classified in- and out-of-distribution samples for the \textbf{MahaAD} approach. The examples confirm that \textbf{MahaAD} typically classifies obvious in-distribution or OoD samples correctly but can misclassify borderline samples. For instance, some false negatives may be considered in-distribution based on their retinal-like structure, while false positives often exhibit a hyperreflective retinal pigment epithelium layer, which might lead to their OoD classification.

\section{Discussion and conclusion}
In this work, we showed how corrupted data from an iiOCT probe in the context of retinal microsurgery can be rejected from further evaluation by using unsupervised OoD detection. The simple MahaAD approach was able to maintain good performance of distance estimation by reliably detecting and rejecting simulated corruptions, and showed promising results on OoD cases from an ex vivo porcine trial. 

The experiments revealed that the benefits of MahaAD observed for a variety of scenarios on 2D~images~\cite{Doorenbos2022} translate well to temporal iiOCT scans with high levels of noise and limited lateral view. Another benefit is its computational efficiency, allowing it to cope with high-frequency A-scan acquisition with minimal latency. Additionally, the experiments point to the challenges of supervised OoD detection when not all unknowns (\ie, possible corruptions) are known and why unsupervised OoD detection might be suitable for improved generalization. In conclusion, we showed that detecting corrupted iiOCT data through unsupervised OoD detection is feasible and that MahaAD could potentially be used to improve safety in retinal microsurgery.

However, one limitation of this work is that the temporal component of the iiOCT is largely ignored as individual samples were considered {for the distance estimation} without any knowledge of the past. In the future, we plan to take this temporal information into account by combining the MahaAD OoD detection with dedicated techniques such as Bayesian filters to further improve performances.

\backmatter

\section*{Declarations}


\begin{itemize}
\item Funding: This work was supported by EUREKA Eurostars (project \#114442) and H2020 project GEYEDANCE. 
\item Competing interests: The authors have no conflict of interest.
\item Ethics approval: Not applicable.
\item Consent: Not applicable.
\item Availability of data, materials, and code: Data is private, code and models are available at \url{https://github.com/alainjungo/ipcai23-iioct-ood}.  
\end{itemize}

\end{document}